\documentclass[conference]{IEEEtran}
\IEEEoverridecommandlockouts
\usepackage{lscape}
\usepackage{rotating}
\usepackage{epstopdf}
\usepackage{etoolbox}
\usepackage{nomencl}
\usepackage{supertabular}
\usepackage{booktabs}
\usepackage{amsfonts}
\usepackage[dvips]{epsfig}
\usepackage{latexsym}
\usepackage{amssymb}
\usepackage{amsmath}
\usepackage{verbatim}
\usepackage{setspace}
\usepackage[ruled,linesnumbered,boxed,commentsnumbered]{algorithm2e}
\usepackage{bm}
\usepackage[colorlinks=true, linkcolor=black, citecolor=blue, urlcolor=blue]{hyperref}       
\usepackage{url}            
\usepackage{lipsum}	

\usepackage{cite}
\usepackage{amsmath,amssymb,amsfonts}
\usepackage{algorithmic}
\usepackage{graphicx}
\usepackage{picinpar}
\usepackage[utf8]{inputenc}
\usepackage{colortbl}
\usepackage{soul}
\usepackage{pifont}
\usepackage{alltt}
\usepackage{enumerate}
\usepackage{breakurl}
\usepackage{subfigure}
\usepackage{epstopdf}
\usepackage{pbox}
\usepackage{fancyhdr}

\usepackage{stfloats}
\usepackage{url}

\usepackage[ruled,linesnumbered,boxed,commentsnumbered]{algorithm2e}
\usepackage[dvips]{epsfig}
\usepackage{lineno}
\modulolinenumbers[5]
\usepackage{multirow}
\usepackage{comment}

\usepackage{xcolor}
\usepackage{underscore}
\usepackage{optidef}
\def\BibTeX{{\rm B\kern-.05em{\sc i\kern-.025em b}\kern-.08em
    T\kern-.1667em\lower.7ex\hbox{E}\kern-.125emX}}
\usepackage[figure]{hypcap}
\hypersetup{hypertex=true,
            colorlinks=true,
            linkcolor=blue,
            anchorcolor=blue,
            citecolor=blue}

\begin{document}

\title{Seabird-Aware Multi-Objective Offshore Wind Capacity Allocation: A Gippsland Case Study}

\author{\IEEEauthorblockN{Leilah Taouk\textsuperscript{1}, Canchen Jiang\textsuperscript{1*}, Markus Wagner\textsuperscript{1}, Yuqing Xiao\textsuperscript{1}, Yidan Zhang\textsuperscript{2}}
\IEEEauthorblockA{\textsuperscript{1} Department of Data Science and AI, Faculty of IT, Monash University, Australia \\
\textsuperscript{2} Department of Human-Centred Computing, Faculty of IT, Monash University, Australia\\
}
\thanks{*Corresponding author: Canchen Jiang (canchen.jiang1@monash.edu).}
}

\maketitle

\begin{center}
\footnotesize
\textit{This work has been submitted to the IEEE for
possible publication. Copyright may be transferred
without notice, after which this version may no
longer be accessible.}
\end{center}

\begin{abstract}
Offshore wind planning requires balancing capital cost, wind-resource potential, and seabird exposure. We formulate a spatial tri-objective mixed-integer linear program for offshore wind capacity allocation, minimizing capital cost and modeled seabird exposure while maximizing wind-weighted capacity. Applied to Australia's Gippsland offshore wind area, the model identifies exposure reductions of 23-44\% at capital-cost premiums of 0.8-6.1\% across 15 wind-weighted capacity requirements. We benchmark exposure-aware optimization against 19 percentile-based exclusion thresholds. At a fixed wind-weighted capacity requirement, the 5th-percentile threshold reduces exposure by 30.1\% at a 37.6\% cost premium, versus at least 30.1\% at a 1.4\% premium with optimization. Under a fixed capital budget, exclusion screening reduces exposure by 48.2\% with a 26.9\% capacity loss, compared with 48.5\% and 7.6\%, respectively, under optimization. These results highlight the value of explicitly optimizing graded seabird exposure.
\end{abstract}

\section{Introduction} \label{sec:intro}
Offshore wind energy plays an important role in electricity-system decarbonization, supported by favorable wind resources and the potential for large-scale electricity generation~\cite{diaz2020review}. Realizing this potential requires careful site selection, as wind resources, development costs, and environmental sensitivities vary spatially. Offshore wind planning therefore involves trade-offs among energy potential, economic costs, and environmental impacts.

One important environmental concern is the potential impact of offshore wind development on seabirds, including collision, displacement, and barrier effects~\cite{reidImpactsBirdsOffshore2022}. Because seabird distributions vary spatially, the location of offshore wind development can influence seabird exposure.

A common approach to addressing environmental concerns in offshore wind planning is spatial exclusion, whereby protected or ecologically sensitive areas are removed from development consideration~\cite{cranmerPortfolioModelSiting2018a}. Seabird sensitivity maps can inform such screening~\cite{bradburyMappingSeabirdSensitivity2014}, but applying a threshold converts graded sensitivity information into a binary distinction between eligible and excluded sites. Although exclusions are necessary where development is prohibited, threshold-based screening alone does not account for differences in exposure among eligible sites when allocating capacity. It may therefore overlook alternative capacity allocations with lower modeled seabird exposure.

Beyond determining site eligibility, offshore wind planning requires decisions on where and how much capacity to develop. Optimization-based approaches have been applied to support these decisions, including multi-objective methods for evaluating competing techno-economic objectives in wind-farm site selection~\cite{mytilinou2017multi}. Mixed-integer formulations have also been developed to select portfolios of offshore wind sites while accounting for economic performance and wake interactions between farms~\cite{cranmerPortfolioModelSiting2018a}. 

Other studies have explicitly examined trade-offs between ecological considerations and offshore wind development. Best and Halpin~\cite{bestMinimizingWildlifeImpacts2019} combined wind-farm profitability and seabird sensitivity using a weighted utility measure. White et al.~\cite{whiteSpatialPlanningOffshore2024} and Wallach et al.~\cite{wallach2025seabirds} examined trade-offs between wind-energy development and ecological considerations without separately optimizing project capital costs. Virtanen et al.~\cite{virtanen2022balancing} integrated economic, ecological, and societal factors into offshore wind spatial prioritization, while Ferguson et al.~\cite{fergusonFlexibleFrameworkSpeciesbased2025} proposed a species-based cumulative-effects framework that uses spatial optimization to minimize ecological impacts while meeting a specified development target.


Despite these advances, the economic and wind-resource benefits of explicitly optimizing graded seabird exposure, relative to threshold-based exclusion screening under equivalent development requirements, remain insufficiently quantified.

To address this gap, we develop a spatially explicit tri-objective mixed-integer linear program (MILP) that minimizes capital cost and modeled seabird exposure while maximizing wind-weighted capacity, a proxy for energy-generation potential based on installed capacity and site-level wind resources. The augmented epsilon-constraint method (AUGMECON2)~\cite{mavrotasImprovedVersionAugmented2013} is employed to approximate the Pareto-optimal set and investigate the trade-offs among these objectives. We further compare exposure-aware optimization with percentile-based exclusion screening under equivalent planning requirements.

The main contributions of this paper are summarized as follows:
\begin{itemize}
    \item We formulate a spatial multi-objective MILP for offshore wind capacity allocation that explicitly considers capital cost, wind-weighted capacity, and modeled seabird exposure as separate objectives.
    \item We systematically compare exposure-aware optimization with 19 percentile-based exclusion thresholds under fixed wind-weighted capacity requirements and fixed capital budgets, evaluating the economic and energy trade-offs associated with reducing seabird exposure.
    \item Through a case study in the Gippsland offshore wind area, Australia, we demonstrate that exposure-aware optimization achieves comparable seabird exposure reductions of 30-33\% at capital-cost premiums below 1.5\%, compared with 21.2-37.6\% under restrictive exclusion screening. Under a fixed capital budget, it achieves approximately 48\% exposure reduction with a 7.6\% loss of wind-weighted capacity.
\end{itemize}

\section{Multi-Objective Offshore Wind Planning Formulation}
\label{sec:formulation}
\subsection{Planning Decisions and Constraints}
The offshore study region is discretized into 1~km $\times$ 1~km spatial cells, with geographic and technical exclusions applied to define the set of eligible candidate cells $\mathcal{C}$. For each cell $w\in\mathcal{C}$, the binary variable $b_w\in\{0,1\}$ indicates whether the cell is selected for development, and $t_w$ denotes its allocated installed capacity (MW). Note that, in this work, the model allocates capacity across cells rather than determining individual turbine locations. The allocation is subject to the following constraints:
\begin{subequations}
\begin{align}
& \underline{t}b_w \leq t_w\leq\bar{t}b_w, \qquad \forall w\in\mathcal{C}.
\label{eq:planning_constraints} \\
& Q^{\min} \le \sum_{w\in\mathcal{C}}q_wt_w \le Q^{\max},
 \label{eq:wind_requirement}
\end{align}
\end{subequations}
where $\underline{t}$ and $\bar{t}$ denote the minimum and maximum installed capacities per developed cell, respectively. $Q^{\min}$ and $Q^{\max}$ denote the minimum and maximum
wind-weighted capacity bounds (MW). Since these limits apply to aggregate wind-weighted output rather than nameplate capacity, total installed capacity may exceed $Q^{\max}$. The wind-power coefficient $q_w\in[0,1]$ is defined in
Section~\ref{wind-energy}.


\subsection{Objective functions}
The proposed model considers three objectives: minimizing capital cost and modeled seabird exposure while maximizing wind-weighted capacity. The corresponding objective functions are defined as follows.

\subsubsection{Capital Cost}
The total capital cost is expressed as
\begin{equation}
\mathbf{F_1} =
\sum_{w\in\mathcal{C}} u_w t_w,
\label{eq:objective_cost}
\end{equation}
where $u_w$ is the estimated capital cost per MW in cell $w$, calculated as $u_w = 1000\left[ C^{\mathrm{cap}} \left(1+\phi\left(\frac{h_w}{\bar h}-1\right)\right) + C^{\mathrm{wire}}L_w \right]$. Here, $C^{\mathrm{cap}}$ denotes the reference capital cost (\$/kW), adjusted for local water depth $h_w$ relative to the mean depth $\bar{h}$. The parameter $\phi$ represents the foundation share of capital cost, assuming a linear relationship between foundation cost and water depth. The connection cost is approximated using the export-cable cost $C^{\mathrm{wire}}$ (\$/kW/km), and straight-line distance $L_w$ to the onshore connection point. $\mathbf{F_1}$ represents capital expenditure only; operating and maintenance costs are outside the model scope. All costs are expressed in AUD.

\subsubsection{Seabird Exposure}
Aggregate seabird exposure across selected cells is calculated as
\begin{equation}
\mathbf{F_2} =
\sum_{w\in\mathcal{C}} e_w b_w,
\label{eq:objective_exposure}
\end{equation}
where $e_w$ is the multi-species exposure score for cell $w$, derived from survey observations of tube-nosed seabirds (Procellariiformes) \cite{BirdlifeData}. Species-level exposure estimates are combined using a weighted average:
$
e_w =
\frac{\sum_{s\in\mathcal{S}}\omega_s\hat d_{sw}}
{\sum_{s\in\mathcal{S}}\omega_s},
$
where $\hat d_{sw}\in[0,1]$ is the estimated detection probability for species $s$ in cell $w$, and $\omega_s$ is its assigned weight. Equal species weights are used in this study. The objective therefore measures cumulative exposure across developed cells, with each cell contributing its exposure score regardless of installed capacity.

\subsubsection{Wind-Weighted Capacity} \label{wind-energy}
Wind-weighted installed capacity, serving as a proxy for energy generation potential, is given by
\begin{equation}
\mathbf{F_3} = \sum_{w\in\mathcal{C}}q_w t_w,
\label{eq:objective_energy}
\end{equation}
where $q_w\in[0,1]$ is the wind-power coefficient for cell $w$, calculated from its mean hub-height wind speed $\bar v_w$ using a piecewise approximation of the turbine power curve~\cite{IEArefTurbine}:
\begin{equation}
q_w =
\begin{cases}
0, & \bar v_w<v_{\mathrm{in}},\\[2pt]
\dfrac{\bar v_w^3-v_{\mathrm{in}}^3}
{v_r^3-v_{\mathrm{in}}^3},
& v_{\mathrm{in}}\leq\bar v_w<v_r,\\[6pt]
1, & v_r\leq\bar v_w\leq v_{\mathrm{out}},\\[2pt]
0, & \bar v_w>v_{\mathrm{out}},
\end{cases}
\label{eq:wind_coefficient}
\end{equation}
where $v_{\mathrm{in}}$, $v_r$, and $v_{\mathrm{out}}$ denote the turbine cut-in, rated, and cut-out speeds, respectively. The coefficient $q_w$ approximates turbine output at the mean wind speed as a proportion of rated power. Because only annual mean wind speeds are available, $\mathbf{F_3}$ represents wind-weighted capacity (MW) rather than estimated annual energy production.

\subsection{Multi-Objective Formulation}
The tri-objective optimization is formulated as
\begin{equation}
\begin{aligned}
& \min_{b_w,t_w, \forall w \in \mathcal{C}}\quad \left\{\mathbf{F_1}, \mathbf{F_2}, -\mathbf{F_3} \right\}\ \\
& \mathrm{s.t.}\quad
\eqref{eq:planning_constraints}-\eqref{eq:wind_requirement}.
\end{aligned}
\label{eq:multiobjective_problem}
\end{equation}

\section{Solution Method}
The multi-objective problem is solved using the augmented $\varepsilon$-constraint method (AUGMECON2)~\cite{mavrotasImprovedVersionAugmented2013} to generate an approximation of the Pareto-optimal set. Capital cost $\mathbf{F_1}$ is retained as the primary objective, while modeled seabird exposure $\mathbf{F_2}$ and wind-weighted capacity $\mathbf{F_3}$ are bounded by systematically varied $\varepsilon$-constraints to generate alternative trade-off solutions.

For each pair of $\varepsilon$-levels $(\varepsilon_2,\varepsilon_3)$, the following augmented single-objective problem is solved by
\begin{equation}
\begin{aligned}
\min_{b_w, t_w, s_2,s_3}\quad
&
\frac{\mathbf{F_1}-z_1^*}{R_1} -\gamma\left(\frac{s_2}{R_2} +0.1\frac{s_3}{R_3} \right)\\
\mathrm{s.t.}\quad
& \mathbf{F_2}+s_2=\varepsilon_2, \mathbf{F_3}-s_3=\varepsilon_3,\\
&
\eqref{eq:planning_constraints}-\eqref{eq:wind_requirement},  
\end{aligned}
\label{eq:augmecon}
\end{equation}
where $z_1^*$ is the ideal value of $F_1$, $R_i$ denotes the payoff-table range of objective $i$, and $\gamma$ is a small augmentation coefficient. The $\varepsilon$-levels impose an upper bound on seabird exposure ($\varepsilon_2$) and a lower bound on wind-weighted capacity ($\varepsilon_3$), and $s_2$, $s_3$ are corresponding non-negative slack variables. The augmentation term rewards improvements in the constrained objectives beyond their prescribed bounds, discouraging weakly Pareto-optimal solutions.

\subsection{Grid Search and Implementation}
To generate the $\varepsilon$-constraint subproblems, a payoff table is first constructed through lexicographic optimization of the three objectives. This establishes the search ranges $\varepsilon_2\in[1.75,25.54]$ and $\varepsilon_3\in[300,2{,}210]$~MW. Each range is discretized into 16 levels, yielding up to 256 subproblems.

The augmentation coefficient is set to $\gamma=10^{-4}$, with a factor of $0.1$ applied to the wind-weighted capacity slack following the AUGMECON2 weighting convention~\cite{mavrotasImprovedVersionAugmented2013}. All subproblems are solved using Gurobi with a relative optimality gap of $10^{-4}$. To improve computational efficiency, infeasible subproblems are skipped where tighter $\varepsilon$-constraints guarantee infeasibility, and successive subproblems are warm-started. Duplicate and dominated solutions are then removed to obtain the final approximation of the Pareto-optimal set.

\section{Gippsland Case Study}
\subsection{Simulation Settings}
The study area is the Gippsland declared offshore wind area (approximately 15,000 km$^2$), discretized into 1~km $\times$ 1~km cells. Cells deeper than 80~m or within 3 nautical miles (5.556~km) of the coastline are excluded, leaving 7,406 candidate cells.

Bathymetry is obtained from the AusBathyTopo 2024 1-km grid~\cite{beaman2024ausbathytopo}, and wind conditions are represented by 2024 annual mean wind speeds at 150~m hub height. Wind-power coefficients are calculated using the IEA 15-MW reference turbine with cut-in, rated, and cut-out speeds of 3.0, 10.59, and 25.0~m\,s$^{-1}$, respectively\cite{IEArefTurbine} (Section~\ref{wind-energy}).

Seabird exposure is derived from at-sea survey observations of tube-nosed seabirds (Procellariiformes). Ordinary kriging is used to interpolate species-level detection probabilities onto the planning grid, yielding both cell-level exposure scores $e_w$ and prediction standard deviations $\hat{\sigma}_{e_w}$.

The reference capital cost is $C^{\mathrm{cap}}=5{,}433$AUD/kW, with a foundation-cost fraction of $\phi=0.128$\cite{CSIROgencost}. Export-cable cost is $C^{\mathrm{wire}}=4.98$AUD/kW/km\cite{gaoDetailedMappingTechnical2024}, with Loy Yang as the grid connection point. Installed capacities per cell are $\underline{t}=3.3$ and $\bar{t}=20.2$ MW, respectively, corresponding to the assumed development density of 3.3–20.2 MW/km$^{2}$~\cite{enevoldsenDataInvestigationInstalled2021}. The wind-weighted capacity bounds are set to $Q^{\min}=300$~MW and $Q^{\max}=2{,}210$~MW.

\subsection{Trade-offs across the three objectives}
\begin{figure}[h]
\centering
\includegraphics[width=\linewidth]{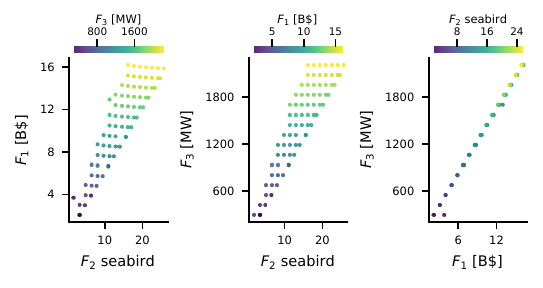}
\caption{Pairwise projections of the 178 non-dominated solutions. Left: $\mathbf{F_2}$ (seabird exposure, $x$) vs $\mathbf{F_1}$ (capital cost, $y$), colored by $\mathbf{F_3}$. Center: $\mathbf{F_2}$ vs $\mathbf{F_3}$, colored by $\mathbf{F_1}$. Right: $\mathbf{F_1}$ vs $\mathbf{F_3}$, colored by $\mathbf{F_2}$.}
\label{fig:pareto}
\end{figure}

The optimization produced 178 non-dominated solutions, spanning capital costs of B\$2.12--16.14, seabird-exposure scores of 1.75-25.51, and wind-weighted capacities of 300--2,210~MW. Fig.~\ref{fig:pareto} shows all three pairwise projections of the Pareto front.

\begin{figure}[h]
\includegraphics[width=\linewidth]{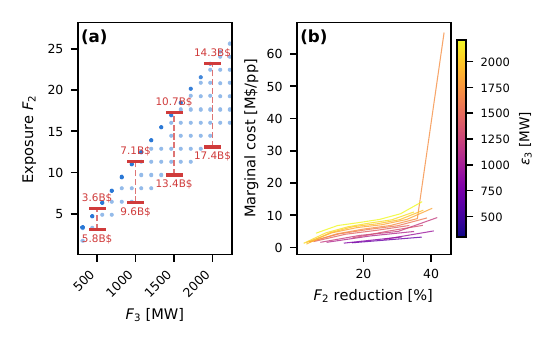}
\caption{(a)~Seabird-exposure variation among non-dominated solutions at four capacity reference levels ($F_3\approx500$, 1000, 1500, 2000~MW). Red bars mark the minimum and maximum $\mathbf{F_2}$ within $\pm250$~MW of each reference; annotations give $\mathbf{F_1}$ at the exposure extremes.
(b)~One curve per capacity level ($\varepsilon_3=427$–$2{,}210$~MW; 15 levels total, lighter = higher wind-weighted requirement): marginal capital cost of seabird-exposure reduction [M\$ per percentage-point reduction in $F_2$] versus cumulative $F_2$ reduction from the minimum-cost baseline.}
\label{fig:BJ}
\end{figure}

Capital cost and seabird exposure both increase strongly with wind-weighted capacity (Spearman's $\rho=0.997$ and $0.956$, respectively). Nevertheless, at four reference capacity levels (500--2{,}000~MW), the highest-exposure solution carries 2.0--4.5$\times$ more exposure than the lowest-exposure solution at comparable capacity (Fig.~\ref{fig:BJ}(a)). This variation highlights the importance of spatial allocation in minimizing seabird exposure.

\subsection{Cost of reducing seabird exposure}
At the lowest wind-weighted capacity requirement ($\varepsilon_3=300$~MW), reducing seabird exposure by 48.1\% increases capital cost by 76.2\% (B\$2.13 to B\$3.74). The minimum-cost and minimum-exposure solutions share no selected cells, and the latter requires 93.7\% more installed capacity to meet the same requirement. Across the remaining 15 capacity levels ($\varepsilon_3 = 427$–$2{,}210$~MW), the largest observed exposure reduction in the sampled set ranges from 23 to 44\% per level, while the corresponding cost increase ranges from 0.8 to 6.1\%, indicating that substantial seabird exposure reductions can be achieved at modest additional cost (Fig.~\ref{fig:BJ}(b)).

Each additional unit of exposure reduction is more costly than the last (Fig.~\ref{fig:BJ}(b)). Initial reductions of 10--20\% are achievable at a cost premium below 1\%, while the final increments incur steeply
rising costs. This empirical pattern is consistent across all 15
capacity levels shown.

\subsection{Spatial patterns of development}

\begin{figure}[h]
\centering
\includegraphics[width=\linewidth]{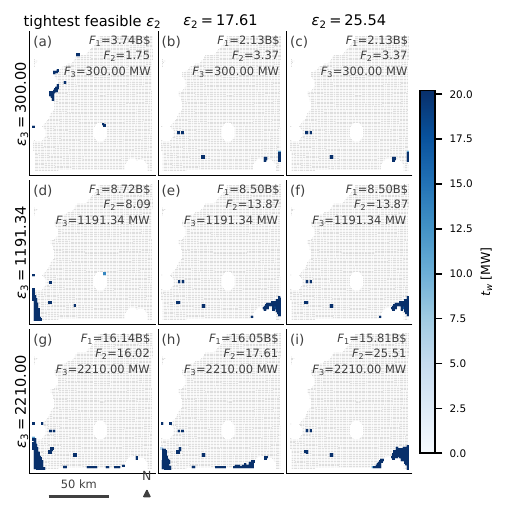}
\caption{Spatial development patterns for nine representative solutions from the full $16{\times}16$ $\varepsilon$-grid. Rows span minimum to maximum wind-weighted capacity floor. The left column shows the tightest feasible seabird-exposure cap at each level, center and right show fixed caps feasible across all rows. Color indicates installed capacity $t_w$.}
\label{fig:epsilon_layouts}
\end{figure}

Fig.~\ref{fig:epsilon_layouts} illustrates nine representative spatial allocations under three wind-weighted capacity requirements and seabird-exposure limits. Higher capacity requirements lead to more extensive development, while tighter exposure limits generally favor lower-exposure locations, particularly toward the southwestern part of this study area. 

The effect of exposure limits varies with the capacity requirement. At $\varepsilon_3=300$ and $1{,}191.3$~MW, relaxing $\varepsilon_2$ from 17.6 to 25.5 does not change the optimal allocation, indicating that these exposure limits are nonbinding. At $\varepsilon_3=2{,}210$~MW, tightening the exposure limit reduces $F_2$ from 25.5 to 16.0, while increasing capital cost from AUD~15.8 to 16.1 billion. These results illustrate how explicit exposure constraints influence the spatial distribution of development and the associated economic trade-offs.

\subsection{Comparison with Exclusion-Based Screening}
We compare exposure-aware optimization against exclusion-based screening across 19 thresholds (5th-95th percentiles), excluding cells above each threshold. Both approaches minimize capital cost at a wind-weighted capacity requirement of 1,319~MW or maximize wind-weighted capacity within a budget of AUD~9.42~billion, the minimum cost of meeting that requirement. Exposure reductions and cost increases are relative to the minimum-cost allocation. Capacity losses are relative to the maximum achievable within the AUD~9.42~billion budget without an exposure limit.

\begin{figure}[h]
\centering
\includegraphics[width=\linewidth]{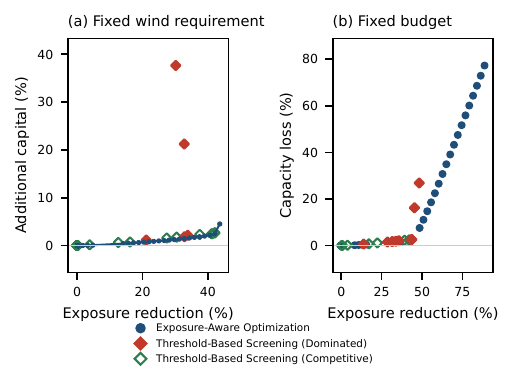}
\caption{Comparison of exposure-aware optimization against exclusion-based screening across 19 percentile thresholds. (a)~Fixed wind-weighted capacity requirement; seabird exposure reduction and capital cost premium relative to the minimum-cost baseline. (b)~Fixed capital budget; seabird exposure reduction and wind-weighted capacity loss relative to the no-exposure-limit baseline. Filled red diamonds indicate screening solutions dominated by at least one sampled optimized allocation in the respective two-objective comparison; open green diamonds have no dominating counterpart in that sample.}
\label{fig:exclusion}
\end{figure}

Screening can produce competitive allocations (Fig.~\ref{fig:exclusion}(a)). The 15th-percentile threshold reduces exposure by 42.1\% for a 2.7\% cost increase and is not dominated by sampled optimized allocations. However, the stricter 5th-percentile threshold increases cost by 37.6\% while reducing exposure by 30.1\%. Optimization achieves at least this reduction with a 1.4\% cost increase.

At a fixed budget (Fig.~\ref{fig:exclusion}(b)), the 5th-percentile threshold reduces exposure by 48.2\% but sacrifices 26.9\% of wind-weighted capacity. Optimization achieves a slightly greater reduction of 48.5\% with only 7.6\% capacity loss.

In this case, restrictive screening limits access to stronger wind resources. Explicitly optimizing seabird exposure therefore achieves comparable or greater reductions with substantially smaller investment or capacity penalties than the most restrictive screening policies. Across all five wind-weighted capacity requirements
(555-2,083 MW), exposure-aware optimization dominates between one and seven screening solutions in the cost-exposure comparison, indicating that its advantages extend beyond the reference case.

\subsection{Sensitivity Analysis}
\label{sec:sensitivity}
Three model inputs carry meaningful uncertainty: kriging-based exposure estimates $e_w$, the foundation-cost fraction $\phi$, and the export-cable cost $C^{\mathrm{wire}}$. We assess the sensitivity of the minimum-cost and minimum-exposure solutions to variation in each input. For exposure, we substitute a one-standard-deviation upward perturbation $\tilde{e}_w = e_w + \hat{\sigma}_{e_w}$ (+25\% of mean $e_w$). For cost, we vary $\phi$ and $C^{\mathrm{wire}}$ each by $\pm30\%$. For each perturbation, we re-optimize using the perturbed inputs. Table~\ref{tab:sensitivity} reports $\Delta F_1$ and $\Delta F_2$
at both the minimum-cost and minimum-exposure operating points.

\begin{table}[ht]
\caption{Sensitivity at $\varepsilon_3=1{,}319$~MW for the minimum-cost (MC)
         and minimum-exposure (ME) operating points.
         $\Delta$ values are relative to the baseline; $F_3=\varepsilon_3$ throughout.}
\label{tab:sensitivity}
\centering
\setlength{\tabcolsep}{4pt}
\footnotesize
\begin{tabular}{l rr rr}
\hline
& \multicolumn{2}{c}{MC point} & \multicolumn{2}{c}{ME point}\\
& \multicolumn{2}{c}{\footnotesize$F_1{=}9.42$~B\$,\ $F_2{=}15.49$}
& \multicolumn{2}{c}{\footnotesize$F_1{=}12.12$~B\$,\ $F_2{=}8.48$}\\
\cline{2-3}\cline{4-5}
Case & $\Delta F_1$ (\%) & $\Delta F_2$ (\%)
     & $\Delta F_1$ (\%) & $\Delta F_2$ (\%)\\
\hline
Kriging $+1\hat{\sigma}$           & $0.0$  & $+27.2$ & $-2.6$ & $+41.3$ \\
$\phi$ $+30\%$                     & $-0.2$ & $+0.2$  & $-0.5$ & $0.0$   \\
$\phi$ $-30\%$                     & $+0.1$ & $-1.1$  & $+0.5$ & $0.0$   \\
$C^{\mathrm{wire}}$ $+30\%$        & $+3.5$ & $-0.5$  & $+2.5$ & $0.0$   \\
$C^{\mathrm{wire}}$ $-30\%$        & $-3.5$ & $+1.0$  & $-2.5$ & $0.0$   \\
\hline
\end{tabular}
\end{table}
Varying $\phi$ and $C^{\mathrm{wire}}$ by $\pm30\%$ shifts $F_1$ by at most 3.5\%, while $F_2$ remains unchanged at the minimum-exposure point and varies by at most 1.1\% at the minimum-cost point. These solutions indicate limited sensitivity to cost assumptions. Under the conservative kriging surface, $F_2$ rises 27.2\% at the minimum-cost point and 41.3\% at the minimum-exposure point, reflecting higher cell-level exposure scores, as well as changes in optimized allocations. Across the tested perturbations, the direction of the cost--exposure trade-off persists, although its magnitude is sensitive to the exposure surface.

\section{Conclusions and Future Work}
\label{sec:conclusion}

This paper developed a tri-objective MILP for offshore wind capacity allocation considering capital cost, modeled seabird exposure, and wind-weighted capacity. In the Gippsland case study, exposure-aware optimization achieved at least a 30.1\% exposure reduction at a 1.4\% capital-cost premium, compared with 37.6\% under 5th-percentile exclusion screening at a fixed wind-weighted capacity requirement. Under a fixed capital budget, it achieved a 48.5\% exposure reduction with a 7.6\% capacity loss, compared with 48.2\% and 26.9\%, respectively, under screening. These results highlight the benefits of directly optimizing graded seabird exposure. Future work will incorporate seasonal seabird distributions and more detailed wake and grid-connection models.

\bibliographystyle{IEEEtran}
\bibliography{ref.bib}

@article{bestMinimizingWildlifeImpacts2019,
  title = {Minimizing Wildlife Impacts for Offshore Wind Energy Development: Winning Tradeoffs for Seabirds in Space and Cetaceans in Time},
  author = {Best, Benjamin D. and Halpin, Patrick N.},
  year = 2019,
  journal = {PLoS One},
  volume = {14},
  number = {5},
  pages = {e0215722},
  publisher = {Public Library of Science, San Francisco, CA USA},
  isbn = {1932-6203}
}

@article{bradburyMappingSeabirdSensitivity2014,
  title = {Mapping Seabird Sensitivity to Offshore Wind Farms},
  author = {Bradbury, Gareth and Trinder, Mark and Furness, Bob and Banks, Alex N. and Caldow, Richard WG and Hume, Duncan},
  year = 2014,
  journal = {PloS one},
  volume = {9},
  number = {9},
  pages = {e106366},
  publisher = {Public Library of Science, San Francisco, USA},
  isbn = {1932-6203}
}

@article{cranmerPortfolioModelSiting2018a,
  title = {A Portfolio Model for Siting Offshore Wind Farms with Economic and Environmental Objectives},
  author = {Cranmer, Alexana and Baker, Erin and Liesi{\"o}, Juuso and Salo, Ahti},
  year = 2018,
  journal = {European Journal of Operational Research},
  volume = {267},
  number = {1},
  pages = {304--314},
  publisher = {Elsevier},
  isbn = {0377-2217}
}

@techreport{CSIROgencost,
  type = {Final {{Report}}},
  title = {{{GenCost}} 2025--26: {{Final}} Report},
  author = {Graham, Paul and Hayward, Jenny and Foster, James},
  year = 2026,
  address = {Australia},
  institution = {CSIRO}
}

@article{gaoDetailedMappingTechnical2024,
  title = {Detailed Mapping of Technical Capacities and Economics Potential of Offshore Wind Energy: {{A}} Case Study in {{South}}-Eastern {{Australia}}},
  author = {Gao, Qiang and Hayward, Jennifer A. and Sergiienko, Nataliia and Khan, Salman Saeed and Hemer, Mark and Ertugrul, Nesimi and Ding, Boyin},
  year = 2024,
  journal = {Renewable and Sustainable Energy Reviews},
  volume = {189},
  pages = {113872},
  publisher = {Elsevier},
  isbn = {1364-0321}
}

@techreport{IEArefTurbine,
  type = {Technical {{Report}}},
  title = {Definition of the {{IEA}} 15-{{Megawatt Offshore Reference Wind Turbine}}},
  author = {Gaertner, Evan and Rinker, Jennifer and Sethuraman, Latha and Zahle, Fredrik and Anderson, Benjamin and Barter, Garrett and Abbas, Nikhar and Meng, Fanzhong and Bortolotti, Pietro and Skrzypinski, Witold and Scott, George and Feil, Roland and Bredmose, Henrik and Dykes, Katherine and Shields, Matt and Allen, Christopher and Viselli, Anthony},
  year = 2020,
  number = {NREL/TP-5000-75698},
  address = {Golden, CO},
  institution = {National Renewable Energy Laboratory}
}

@book{reidImpactsBirdsOffshore2022,
  title = {Impacts on {{Birds}} from {{Offshore Wind Farms}} in {{Australia}}},
  author = {Reid, Keith and Baker, B. and Woehler, E.},
  year = 2022,
  publisher = {Department of Climate Change, Energy, the Envionment \& Water}
}

@article{whiteSpatialPlanningOffshore2024,
  title = {Spatial Planning Offshore Wind Energy Farms in {{California}} for Mediating Fisheries and Wildlife Conservation Impacts},
  author = {White, Crow and Wang, Yi-Hi and Walter, Ryan K. and Ruttenberg, Benjamin I. and Han, Danny and Newman, Eli and Deyle, Ethan R. and Gopal, Sucharita and Kaufman, Les},
  year = 2024,
  journal = {Environmental Development},
  volume = {51},
  pages = {101005},
  publisher = {Elsevier},
  isbn = {2211-4645}
}

@article{diaz2020review,
  title={Review of the current status, technology and future trends of offshore wind farms},
  author={D{\'\i}az, Hugo and Soares, C Guedes},
  journal={Ocean Engineering},
  volume={209},
  pages={107381},
  year={2020},
  publisher={Elsevier}
}

@misc{BirdlifeData,
  author = {{BirdLife Australia}},
  title  = {{Birdata}},
  url    = {https://birdata.birdlife.org.au/},
  note   = {Accessed: Nov. 12, 2026}
}

@article{mytilinou2017multi,
  title={A multi-objective optimisation approach applied to offshore wind farm location selection},
  author={Mytilinou, Varvara and Kolios, Athanasios J},
  journal={Journal of Ocean Engineering and Marine Energy},
  volume={3},
  number={3},
  pages={265--284},
  year={2017},
  publisher={Springer}
}

@article{enevoldsenDataInvestigationInstalled2021,
  title = {Data Investigation of Installed and Output Power Densities of Onshore and Offshore Wind Turbines Worldwide},
  author = {Enevoldsen, Peter and Jacobson, Mark Z.},
  year = 2021,
  journal = {Energy for Sustainable Development},
  volume = {60},
  pages = {40--51},
  publisher = {Elsevier},
  isbn = {0973-0826}
}

@article{mavrotasImprovedVersionAugmented2013,
  title = {An Improved Version of the Augmented {$\varepsilon$}-Constraint Method ({{AUGMECON2}}) for Finding the Exact Pareto Set in Multi-Objective Integer Programming Problems},
  author = {Mavrotas, George and Florios, Kostas},
  year = 2013,
  month = may,
  journal = {Applied Mathematics and Computation},
  volume = {219},
  number = {18},
  pages = {9652--9669},
  issn = {0096-3003},
  doi = {10.1016/j.amc.2013.03.002}
}

@misc{beaman2024ausbathytopo,
  author       = {Beaman, R.},
  title        = {{AusBathyTopo} ({Australia}) 250m 2024:
                  A national-scale depth model (20240011C)},
  year         = {2024},
  howpublished = {Geoscience Australia},
  note         = {Dataset},
  doi          = {10.26186/150050},
  url          = {https://doi.org/10.26186/150050}
}

@article{fergusonFlexibleFrameworkSpeciesbased2025,
  title = {A Flexible Framework for Species-Based Regional Cumulative Effects Assessments to Support Offshore Wind Energy Planning and Management},
  author = {Ferguson, Megan C. and Williams, Kathryn A. and Goodale, M. Wing and Adams, Evan M. and Knaga, Paul and Kingdon, Katrien and {Avery-Gomm}, Stephanie},
  year = 2025,
  journal = {Environmental Impact Assessment Review},
  volume = {114},
  pages = {107912},
  publisher = {Elsevier},
  isbn = {0195-9255}
}

@article{virtanen2022balancing,
  title={Balancing profitability of energy production, societal impacts and biodiversity in offshore wind farm design},
  author={Virtanen, Elina A and Lappalainen, Juho and Nurmi, Marco and Viitasalo, Markku and Tikanm{\"a}ki, Maria and Heinonen, Jaakko and Atlaskin, Evgeny and Kallasvuo, Meri and Tikkanen, Hannu and Moilanen, Atte},
  journal={Renewable and Sustainable Energy Reviews},
  volume={158},
  pages={112087},
  year={2022},
  publisher={Elsevier}
}

@techreport{wallach2025seabirds,
  author      = {Wallach, Eli and Chamberlin, Charles and Jacobson, Arne and others},
  title       = {Seabirds in {3D}: A Framework to Evaluate Collision Vulnerability With Future Offshore Wind Developments},
  institution = {California Energy Commission},
  number      = {CEC-500-2025-037},
  year        = {2025}
}

\end{document}